\documentclass[11pt]{article}

\usepackage[final]{acl}
\usepackage{subcaption}
\usepackage{algorithm}
\usepackage{algpseudocode} 
\usepackage{times}
\usepackage{latexsym}
\usepackage{amsmath}
\usepackage{amssymb}
\usepackage[T1]{fontenc}
\usepackage[utf8]{inputenc}

\usepackage{microtype}

\usepackage{inconsolata}

\usepackage{graphicx}
\usepackage{booktabs}
\title{Verifier-Free RL for LLMs via Intrinsic Gradient-Norm Reward}

\author{
  Xuexiang Wen\textsuperscript{1} \quad
  Hang Yu\textsuperscript{2} \quad
  Linchao Zhu\textsuperscript{1\dag} \quad
  Gaoang Wang\textsuperscript{1\dag} \\
  \textsuperscript{1}Zhejiang University \quad
  \textsuperscript{2}Ant Group \\
  \texttt{\{xuexiangwen, zhulinchao\}@zju.edu.cn} \quad
  \texttt{gaoangwang@intl.zju.edu.cn}
}

\begin{document}
\maketitle
\renewcommand{\thefootnote}{\textdagger}
\footnotetext{Corresponding authors.}
\renewcommand{\thefootnote}{\arabic{footnote}}
\begin{abstract}
While Reinforcement Learning with Verifiable Rewards (RLVR) has recently emerged as a promising post-training paradigm for Large Language Models (LLMs), its dependency on the gold label or domain-specific verifiers limits its scalability to new tasks and domains. In this work, we propose \textbf{\underline{V}erifier-free \underline{I}ntrinsic \underline{G}radient-N\underline{o}rm \underline{R}eward} (\textbf{VIGOR}), a simple reward that uses only the policy model itself. Given a prompt, VIGOR samples a group of completions and assigns higher within-group rewards to outputs that induce smaller $\ell_2$ norms of the teacher-forced negative log-likelihood gradients under the current parameters. Intuitively, lower gradient norms suggest the completion aligns better with the current policy, serving as an intrinsic preference signal for policy optimization. To make this intrinsic signal practical for RL, we correct the systematic length bias of averaged token-level gradients with a $\sqrt{T}$ scaling, and apply group-wise rank shaping to stabilize reward scales across prompts. Across mathematical reasoning benchmarks, VIGOR outperforms the state-of-the-art Reinforcement Learning from Internal Feedback (RLIF) baseline, and it also exhibits cross-domain transfer to code benchmarks when trained only on math data.  For instance, on Qwen2.5-7B-Base post-trained on MATH, VIGOR improves the average math accuracy by +3.31\% and the average code accuracy by +1.91\% over this baseline, while exhibiting more stable training dynamics. The code is available at \url{https://github.com/ZJUSCL/VIGOR}.

\end{abstract}

\section{Introduction}
\begin{figure}[t]
  \centering
  \includegraphics[width=\columnwidth]{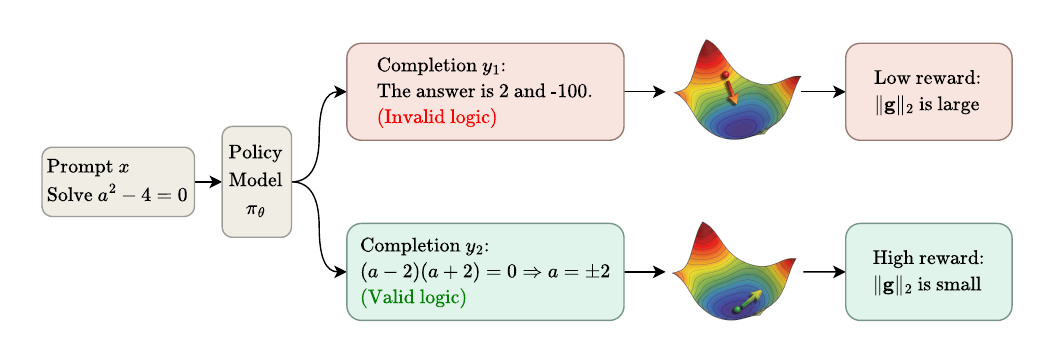}
  \caption{\textbf{Intuition behind VIGOR.}
For a prompt $x$, we sample two completions $y_1, y_2 \sim \pi_\theta$, each inducing its own teacher-forced loss surface $\ell(y_i \mid x; \theta)$ over the parameters $\theta$.
The arrow at each point indicates the direction of the local gradient $\nabla_\theta \ell$, and the steepness of the surface at that point reflects its magnitude $\|\mathbf{g}\|_2$.
Under the current policy, the invalid $y_1$ (red) incurs a higher NLL and sits on a steep slope (large $\|\mathbf{g}\|_2$), while the valid $y_2$ (green) lies in a flatter basin (small $\|\mathbf{g}\|_2$).
VIGOR turns this parameter-space signal into a verifier-free reward by assigning higher reward to completions with smaller $\|\mathbf{g}\|_2$.}

\label{fig:motivation}

  \label{fig:teaser}
\end{figure}
Large Language Models (LLMs) have shown remarkable capabilities in reasoning tasks and general tasks. To further enhance LLMs' performance, Reinforcement Learning with Verifiable Rewards (RLVR) has emerged as a promising post-training paradigm.

However, RLVR depends on the task-specific, programmatic verifiers (e.g., exact-match answer checkers for math or unit tests for code)~\citep{deepseekr1, coevolve}. When such verifiers are unavailable, the reward signal becomes hard to specify, limiting RLVR's applicability and scalability. 

To mitigate this, previous work explores verifier-free alternatives. One crucial line adopts majority voting to derive pseudo-labels from multiple samples, treating the consensus answer as the reward signal~\citep{ttrl, corewarding, majorityvoting}. Despite the promising results, voting-based methods typically assume that final answers can be reliably extracted and normalized for aggregation across samples, which may limit their applicability to open-ended generation settings~\citep{selfconsistency}. Another prominent line constructs intrinsic rewards from the model's internal signals like entropy, enabling verifier-free post-training~\citep{entropy1, intuitor}. Despite the encouraging gains, such intrinsic rewards may become brittle as training progresses, leading to instability and even performance degradation~\citep{nofreelunch}. Overall, we still lack verifier-free rewards that are both broadly applicable to free-form outputs and stable during RL, without relying on auxiliary components beyond the policy model itself.

In this work, we propose \textbf{VIGOR}, a verifier-free intrinsic \emph{gradient-norm} reward computed solely from the policy model. Concretely, for each sampled completion, we compute the gradient norm of the teacher-forced negative log-likelihood under the current parameters (without updating the model), and assign higher  rewards to completions that induce smaller gradient norms. This design is motivated by a first-order optimization view: smaller gradients typically correspond to milder updates and thus smoother training dynamics. To make the signal robust in practice, we (i) correct the systematic length bias with a $\sqrt{T}$ normalization, and (ii) apply rank-based reward shaping to stabilize reward scales across prompts.
We conduct two separate post-training runs on Qwen2.5 base models, one using a mathematics dataset and the other using a code dataset, and evaluate both resulting models on a broad suite of benchmarks spanning mathematical reasoning, code generation, general multi-task capability, and instruction following.

Our contributions are summarized as follows:
\begin{itemize}
  \item We propose VIGOR, a verifier-free intrinsic gradient-norm reward for RL of LLMs; on Qwen2.5-7B post-trained on MATH, it surpasses a state-of-the-art verifier-free RLIF baseline by +3.31 avg.\ math and +1.91 avg.\ code accuracy, with more stable training dynamics.
  \item We introduce $\sqrt{T}$ length correction for gradient norms and rank-based normalization to mitigate length bias and stabilize reward scales across prompts.
  \item We conduct extensive experiments with separate math and code post-training, evaluating both in-domain and cross-domain performance on a broad suite of reasoning, instruction-following, and multi-task benchmarks, with ablations validating both components.
\end{itemize}


\section{Related Work}
\paragraph{Reinforcement Learning with Verifiable rewards.} 
Reinforcement learning from externally verifiable rewards has recently demonstrated strong  improvements in LLMs' reasoning ability on mathematics and programming tasks~\citep{deepseekmath, deepseekr1, dapo,kimik2, mimo}. These works often instantiate an outcome reward model (ORM) that scores only final answers, using PPO-style optimizers such as GRPO \citep{deepseekmath} and DAPO \citep{dapo}. Besides, some researchers focus on the process reward models (PRMs) that provide denser reward signals by assigning step-level feedback along the reasoning trajectory, alleviating the sparsity of final-answer rewards~\citep{letsverifystepstep, stepkto}.

\paragraph{Reinforcement Learning with Intrinsic and Self-Supervised Rewards.} 
Beyond domains with reliable verifiers, obtaining high-quality rewards remains challenging. Recent works therefore explore intrinsic and self-supervised rewards. These rewards are generated by the model itself instead of external verifiers. Methods like TTRL~\citep{ttrl} and Co-rewarding~\citep{corewarding} leverage majority voting~\citep{majorityvoting} to get pseudo labels for optimization. In contrast, some methods exploit the intrinsic model signals, such as self-certainty and entropy-based objectives, to construct the reward without labels~\citep{intuitor,entropy1, entropy2}. EMPO~\citep{empo} also minimizes semantic entropy to improve performance, but its semantic entropy computation relies on auxiliary semantic equivalence model beyond the policy model itself.


\section{Method}
\begin{figure*}[t]
  \centering
  \includegraphics[width=0.85\textwidth]{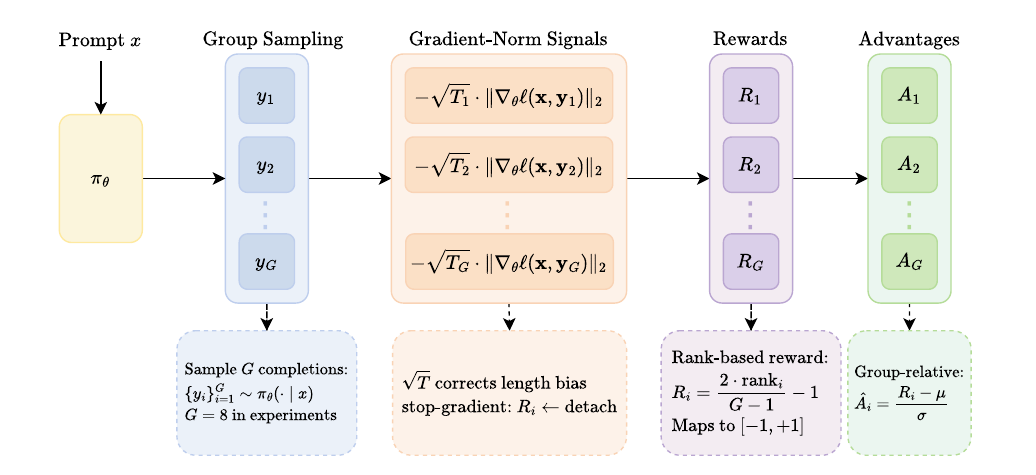}
  \caption{Overview of the proposed method. For each prompt $\mathbf{x}$, we sample a group of $G$ completions from the current policy $\pi_{\theta}$, compute the $\ell_2$ gradient norm of the token-averaged NLL for each completion, and apply a $\sqrt{T}$ length correction to obtain the raw signal $S_{\text{GN}}(\mathbf{x}, \mathbf{y})$. The corrected signals are then mapped to rank-based rewards $R_i \in [-1, +1]$ and normalized into group-relative advantages $\hat{A}_i$ for policy updates.}
  \label{fig:overview}
\end{figure*}
In this section, we propose an intrinsic gradient-norm reward for RL of LLMs.  Unlike RL with verifiable rewards (RLVR), our method does not require an external verifier, and instead leverages internal training dynamics of the language model as supervision. 

\subsection{Preliminaries}
\label{sec:prelim}
We formulate LLM reasoning as a sequence generation problem. Given a prompt $x$, the policy $\pi_\theta$ generates a complete sequence $y=[r;a]$, where $r$ denotes the chain-of-thought reasoning and $a$ denotes the final answer. 

Under this formulation, reinforcement learning (RL) provides a natural framework for optimizing the policy $\pi_\theta$. We review Group Relative Policy Optimization (GRPO) \citep{deepseekmath}, an RL algorithm tailored for LLMs. For each prompt $x$, it samples a group of $G$ completions $\{y_i\}_{i=1}^G$ from $\pi_{\theta}$. The advantage $\hat{A}_i$ for each completion is computed by standardizing the rewards within the group:
\begin{equation}
\hat{A}_i = \frac{R(x, y_i) - \frac{1}{G}\sum_{j=1}^G R(x, y_j)}{\sqrt{\frac{1}{G}\sum_{j=1}^G \left(R(x, y_j) - \bar{R}\right)^2}},
\end{equation}
where $R(x,y_i)$ is the reward for completion $y_i$. The policy model $\pi_{\theta}$ is then updated by maximizing a PPO-style objective that encourages completions with high group-relative advantages $\hat{A}_i$, regularized by the KL divergence between the current and reference policy.

\subsection{Intrinsic Gradient-Norm Reward}
\label{sec:reward}
\paragraph{Motivation.}While GRPO provides an efficient optimization framework, RLVR approaches typically rely on externally verifiable reward signals (e.g., ground-truth matching for math or execution-based signals for code tasks). This reliance constrains their capability to generalize to open-ended or weakly verifiable tasks where such external supervision is unavailable.

From an optimization perspective, first-order stationarity is characterized by near-zero gradients. The gradient norm therefore provides a natural proxy for how close the current parameters are to a stationary region: a smaller norm typically implies weaker first-order variation of the loss locally and consequently more stable parameter updates. Motivated by this observation, we propose an \emph{intrinsic gradient-norm reward} that prefers completions inducing smaller $\ell_2$ gradient norms under the same prompt. Intuitively, such outputs reflect more stable response patterns under the current policy and  assigning them higher relative reward suppresses abrupt gradient fluctuations and encourages smoother, more controllable training dynamics without requiring any external verifier.

\paragraph{The Length Bias Problem.}
For each prompt $x$, we sample a group of completions $\{y_i\}_{i=1}^G$ from 
the current policy $\pi_\theta(x)$ and derive an intrinsic signal from their 
training-dependent gradients. For a completion $y=(y_1,\dots,y_T)$ with 
non-padding length $T=|y|$, we define its average token-level negative 
log-likelihood as
\begin{equation}
\ell_{\text{mean}}(x,y) = \frac{1}{T}\sum_{t=1}^T \ell_t(x,y),
\end{equation}
where $\ell_t(x,y) = -\log \pi_\theta(y_t \mid x, y_{<t})$ denotes the per-token NLL.

We then compute the corresponding gradient 
$\mathbf{g}(x,y) = \nabla_\theta \ell_{\text{mean}}(x,y)$ and treat 
$\|\mathbf{g}(x,y)\|_2$ as an intrinsic signal: smaller norms generally correspond to milder updates and smoother optimization behavior. We detach this gradient norm from the computation graph and use it solely as a scalar reward signal.

However, this choice introduces a systematic length bias: since 
$\ell_{\text{mean}}$ averages $\ell_t$ over $T$ tokens, the gradient 
$\mathbf{g}(x,y)$ is likewise a $1/T$-scaled sum of per-token gradient 
contributions $\nabla_\theta \ell_t(x,y)$. These per-token contributions 
vary in sign and magnitude as the context $y_{<t}$ changes, so as $T$ 
grows they partially cancel across token positions, causing 
$\|\mathbf{g}(x,y)\|_2$ to shrink even when completion quality does not 
improve. As a loose analogy, under an independence assumption on per-token 
contributions, one would expect $\|\mathbf{g}\|_2 = O(1/\sqrt{T})$. 
Although this assumption does not strictly hold for autoregressive LLMs, 
sufficient cancellation still occurs in practice.

\begin{table}[t]
\centering
\small
\begin{tabular}{@{}cccc@{}}
\toprule
Length bin & Avg.\ tokens & $\|\mathbf g\|_2$ & $\sqrt{T}\|\mathbf g\|_2$ \\
\midrule
1 & $\sim 250$  & $\sim 180$ & $\sim 2.85\times 10^3$ \\
2 & $\sim 500$  & $\sim 130$ & $\sim 2.91\times 10^3$ \\
3 & $\sim 750$  & $\sim 105$ & $\sim 2.88\times 10^3$ \\
4 & $\sim 1000$ & $\sim 90$  & $\sim 2.84\times 10^3$ \\
\bottomrule
\end{tabular}
\caption{Empirical verification of $\sqrt{T}$ scaling. We bin training completions by sequence length and report the average raw gradient norm $\|\mathbf g\|_2$ and the rescaled $\sqrt{T}\|\mathbf g\|_2$. The rescaled values remain nearly constant across bins, confirming that $\sqrt{T}$ neutralizes the length-driven variation rather than imposing a length penalty.}
\label{tab:length_bin}
\end{table}

To verify this $O(1/\sqrt{T})$ scaling, we group all training completions into bins by sequence 
length and compute the average gradient norm per bin. As shown in 
Table~\ref{tab:length_bin}, the raw $\|\mathbf{g}\|_2$ varies by roughly 
$2\times$ across length bins, while the rescaled $\sqrt{T}\|\mathbf{g}\|_2$ 
remains nearly constant, confirming the $O(1/\sqrt{T})$ scaling behavior. 
As a result, $\|\mathbf{g}(x,y)\|_2$ tends to shrink as $T$ grows even when 
completion quality does not improve, making a gradient-norm-based reward 
vulnerable to length hacking. To counteract this bias, we multiply by 
$\sqrt{T}$ to produce an approximately length-invariant raw signal:
\begin{equation}
S_{\text{GN}}(x,y) = -\sqrt{T}\,\|\mathbf{g}(x,y)\|_2.
\end{equation}
The negative sign converts gradient-norm minimization into a reward 
maximization objective. This correction is empirically necessary: removing 
$\sqrt{T}$ induces severe length inflation and accuracy collapse 
(Figure~\ref{fig:ablation_fig}), whereas with $\sqrt{T}$ the reward becomes 
approximately length-neutral (Table~\ref{tab:length_bin}).

\paragraph{Rank-based Normalization of the Intrinsic Signal.} 
The raw gradient-norm-based signal $S_{\text{GN}}(x,y)$ is only meaningful in a relative sense within a group of completions, and it may exhibit large scale variations across different prompts. Therefore, we transform the raw signal into a normalized rank-based intrinsic reward. We denote the resulting rank-normalized reward by $R_{\text{GN}}(x,y)$.

For each prompt $x$ and corresponding completions $\{y_i\}_{i=1}^G$, we compute the raw signals $\{S_{\text{GN}}(x,y_i)\}_{i=1}^G$ and sort them from worst to best (smaller $S_{\text{GN}}(x,y_i)$ indicates larger gradient norms and thus worse completions). We assign each completion $y_i$ an integer rank $\text{rank}_i \in \{0,\dots,G-1\}$, with smaller ranks indicating worse completions and larger ranks indicating better ones. The resulting normalized intrinsic reward is given by
\begin{equation}
R_{\text{GN}}(x,y_i)
= 2\,\frac{\text{rank}_i}{G-1}-1.
\end{equation}
This mapping assigns the worst completion ($\text{rank}_i=0$) reward $-1$ and the best completion ($\text{rank}_i=G-1$) reward $+1$, with evenly spaced values in between, making the intrinsic signal comparable across prompts while preserving within-group ordering.

\subsection{Policy Optimization with Intrinsic Gradient-Norm  Reward}

\begin{algorithm}[t]
\caption{Intrinsic Gradient-Norm Reward for GRPO}
\label{alg:vigor}
\small
\begin{algorithmic}[1]
\Require Batch $B$, policy $\pi_\theta$, reference $\pi_{\rm ref}$, group size $G$
\State $\mathcal{T} \gets \emptyset$ \Comment{Initialize trajectory buffer}
\For{\textbf{each} prompt $x \in B$}
    \State \textit{// 1. Sample completions}
    \State Sample group $\{y_i\}_{i=1}^{G} \sim \pi_\theta(\cdot \mid x)$
    
    \State \textit{// 2. Compute gradient norms}
    \For{$i=1$ to $G$}
        \State $g_i \gets \left\lVert \nabla_\theta \ell_{\rm mean}(x,y_i) \right\rVert_2$ \Comment{detach gradient}
        \State $S_i \gets -\sqrt{|y_i|}\, g_i$ \Comment{length correction}
    \EndFor
    
    \State \textit{// 3. Estimate advantage}
    \State $\{R_i\} \gets \textsc{RankNorm}(\{S_i\})$ \Comment{map ranks to $[-1,1]$}
    \State $\{\hat{A}_i\} \gets \textsc{Normalize}(\{R_i\})$ \Comment{$(R_i - \mu) / \sigma$}
    
    \State $\mathcal{T} \gets \mathcal{T} \cup \{(x,y_i,\hat{A}_i)\}_{i=1}^{G}$
\EndFor
\State Update $\theta$ maximizing GRPO objective on $\mathcal{T}$ with KL$(\pi_\theta,\pi_{\rm ref})$
\end{algorithmic}
\end{algorithm}
Let $\mathcal{D}_X$ denote an unlabeled prompt dataset. In this section, we integrate our proposed  intrinsic gradient-norm reward into the GRPO framework. This approach optimizes the policy $\pi_{\theta}$ to favor completions with lower gradient norms given prompts $x \sim \mathcal{D}_{X}$.

\paragraph{Objective Formulation.}
Given a prompt $x$, we sample a group of $G$ completions $\{y_i\}_{i=1}^G \sim \pi_{\theta}(\cdot \mid x)$. We assign the rank-normalized intrinsic reward $R_{\text{GN}}$ to each generation $y_i$. Then our method approximately minimizes the gradient norm $\lVert \mathbf g(x,y) \rVert_{2}$ by maximizing the following objective:
\begin{equation}
\begin{aligned}
\mathcal{J}(\pi_\theta)
= \mathbb{E}\bigg[
&\tfrac{1}{G}\sum_{i=1}^{G}
\tfrac{1}{|y_i|}\sum_{t=1}^{|y_i|}
\min\!\big(r_{i,t}(\theta)\hat{A}_i,\,\bar{r}_{i,t}\hat{A}_i\big) \\
&-\beta\,\mathrm{KL}(\pi_\theta,\pi_{\text{ref}})
\bigg],
\end{aligned}
\end{equation}
where $r_{i,t}(\theta)=\frac{\pi_\theta(y_{i,t}\mid x,\,y_{i,<t})}{\pi_{\theta_{\text{old}}}(y_{i,t}\mid x,\,y_{i,<t})}$ is the token-level probability ratio, $\bar{r}_{i,t}=\mathrm{clip}(r_{i,t}(\theta),\,1{-}\epsilon,\,1{+}\epsilon)$ is its clipped version, and $\hat{A}_i$ is the group-relative advantage computed by mean-std normalizing $\{R_{\text{GN}}(x,y_j)\}_{j=1}^{G}$ within each prompt group.
\paragraph{Stop-gradient Operation.}
Although $R_{\text{GN}}(x,y)$ is computed from the current model parameters via the gradient norm, we treat
$R_{\text{GN}}$ as a constant scalar and detach it from the computation graph during the policy optimization. This avoids backpropagating via the gradient-norm computation and keeps the update first-order.

\begin{table*}[t]
    \centering
    \small
    \setlength{\tabcolsep}{3.5pt}
    \renewcommand{\arraystretch}{1.05}
    \begin{tabular}{@{}lccccccc cc@{}}
        \toprule
         &
        \multicolumn{4}{c}{\textbf{Mathematics}} &
        \multicolumn{3}{c}{\textbf{Code}} &
        \multicolumn{1}{c}{\textbf{Multi-task}} &
        \multicolumn{1}{c}{\textbf{Instruction}} \\
        \cmidrule{1-10}
        \textbf{Methods} &
        \textbf{GSM8K} & \textbf{MATH500} & \textbf{AMC} & \textbf{Avg.} &
        \textbf{LiveCodeBench} & \textbf{CRUX} & \textbf{Avg.} &
        \textbf{MMLU-Pro} & \textbf{IFEval} \\
        \midrule

        \multicolumn{10}{@{}l}{\textbf{Qwen2.5-3B-Base}} \\
        \addlinespace[2pt]
        Before RL (Base)  & 67.93 & 54.80  & 22.28 & 48.34 &  9.57 & 24.38 & 16.98 & 36.92 & 28.30 \\
        - GT-Reward       & 81.96 & 65.60  & 29.97 & 59.17 &  12.32 & 33.00 & 22.66 & 38.17 & 29.91 \\
        - INTUITOR        & 79.68 & 62.40  & 29.21 & 57.10 & 14.88 & 38.70  & 26.79 & 24.48 & 29.11 \\
        - Ours (VIGOR) & \textbf{81.80} & \textbf{64.60} & \textbf{31.02} & \textbf{59.14} &
                           \textbf{15.90} & \textbf{40.00} & \textbf{27.95} &
                           \textbf{32.65} & \textbf{31.72} \\

        \addlinespace[0.5em]
        \multicolumn{10}{@{}l}{\textbf{Qwen2.5-7B-Base}} \\
        \addlinespace[2pt]
        Before RL (Base)  & 43.06 & 63.00    & 21.68 & 42.58 &  1.99 & 17.38 &  9.69 & 47.21 & 35.90 \\
        - GT-Reward       & 84.80 & 74.60  & 42.01 & 67.14 &  7.39 & 55.50  & 31.45 & 43.17 & 34.63 \\
        - INTUITOR        & 87.19 & \textbf{76.20}  & 35.99 & 66.46 & 19.81 & \textbf{57.20} & 38.51 & 43.04 & 34.91 \\
        - Ours (VIGOR) & \textbf{88.70} & \textbf{76.20} & \textbf{44.42} & \textbf{69.77} &
                           \textbf{24.45} & 56.38 & \textbf{40.42} &
                           \textbf{43.09} & \textbf{37.03} \\

        \bottomrule
    \end{tabular}
    \caption{Main results of models trained on the \textbf{MATH} dataset. \textbf{Bold} marks the better score between \textbf{Ours} and \textbf{INTUITOR}. Avg. is the arithmetic mean over tasks in each block.}
    \label{tab:training_on_math}
\end{table*}
\begin{table*}[t]
    \centering
    \small
    \setlength{\tabcolsep}{3.5pt}
    \renewcommand{\arraystretch}{1.05}
    \begin{tabular}{@{}lccccccc cc@{}}
        \toprule
         &
        \multicolumn{4}{c}{\textbf{Mathematics}} &
        \multicolumn{3}{c}{\textbf{Code}} &
        \multicolumn{1}{c}{\textbf{Multi-task}} &
        \multicolumn{1}{c}{\textbf{Instruction}} \\
        \cmidrule{1-10}
        \textbf{Methods} &
        \textbf{GSM8K} & \textbf{MATH500} & \textbf{AMC} & \textbf{Avg.} &
        \textbf{LiveCodeBench} & \textbf{CRUX} & \textbf{Avg.} &
        \textbf{MMLU-Pro} & \textbf{IFEval} \\
        \midrule

        \multicolumn{10}{@{}l}{\textbf{Qwen2.5-3B-Base}} \\
        \addlinespace[2pt]
        Before RL (Base)  & 67.93 & 54.80 & 22.28 & 48.34 &
                           9.57  & 24.38 & 16.98 &
                           36.92 & 28.30 \\
        - INTUITOR        & 75.13 & 58.60 & 22.59 & 52.11 &
                           11.47 & \textbf{39.38} & \textbf{25.43} &
                           29.07 & 29.98 \\
        - Ours (VIGOR)    & \textbf{77.10} & \textbf{62.80} & \textbf{29.82} & \textbf{56.57} &
                           \textbf{11.65} & 35.62 & 23.64 &
                           \textbf{35.01} & \textbf{32.39} \\
        \bottomrule
    \end{tabular}
    \caption{Main results of models trained on the \textbf{CodeContests} dataset. \textbf{Bold} marks the better score between \textbf{Ours} and \textbf{INTUITOR}. Avg. is the arithmetic mean over tasks in each block.}
    \label{tab:training_on_codecontests}
\end{table*}

\section{Experiment}
\subsection{Experiment Setup}
\paragraph{Models.}
We conduct experiments on the Qwen2.5 Base models~\citep{qwen25} for fair comparison with prior work and thorough evaluation across model sizes. Specifically, we use Qwen2.5-3B-Base and Qwen2.5-7B-Base.
\paragraph{Training Data.}
We adopt CodeContests~\citep{alphacode} as the code training dataset and MATH~\citep{math7500} as the mathematics training dataset. For simplicity and training efficiency, we use the first 3,200 problems in CodeContests as the training set. For the MATH dataset, we use the full set of 7,500 training problems.

\paragraph{Evaluation.}
We evaluate our post-trained models on a diverse suite of benchmarks spanning mathematical reasoning, code generation, instruction following, and general multi-task knowledge. For mathematical reasoning, we report results on MATH-500~\citep{math500}, GSM8K~\citep{gsm8k}, and AMC~\citep{numina_math_datasets}. For code reasoning, we use LiveCodeBench (v6)~\citep{livecodebench} and CRUX~\citep{crux}. To assess instruction-following capability, we include IFEval~\citep{ifeval}. Finally, we measure broad multi-task proficiency on MMLU-Pro~\citep{mmlupro}. Unless otherwise specified, we follow the standard evaluation protocol for each benchmark and report pass@1 (or accuracy for multiple-choice benchmarks); for AMC, we additionally use a higher-temperature sampling setting and average results across multiple runs to reduce variance. More evaluation details are provided in Appendix~\ref{sec:additional_evaluation_details}.

\paragraph{Baselines.}
We compare our method with a pretrained reference and two post-training baselines, including RL with verifiable rewards (RLVR) and verifier-free RL from internal feedback (RLIF).
\begin{itemize}
    \item \textbf{Before RL.} The pretrained Qwen2.5-Base model without any parameter updates.
    \item \textbf{GT-Reward.} We reproduce the GRPO baseline implemented in the Open-R1 framework, where the reward is computed by an exact-match verifier against the ground-truth final answer: we extract the final answer from each completion, compare it with the reference, and use the resulting correctness signal as the reward.
    \item \textbf{INTUITOR.} A verifier-free intrinsic-reward baseline under the Reinforcement Learning from Internal Feedback (RLIF) paradigm. INTUITOR replaces the verifiable reward in GRPO with the policy model's own confidence signal computed from its likelihood, enabling fully unsupervised RL without gold labels or domain-specific verifiers~\citep{intuitor}.

\end{itemize}

\paragraph{Implementation Details.}

Both our method and GRPO baseline are trained with Open-R1 framework~\citep{openr1}. For VIGOR training, we discard all reference solutions/labels and use only the problem statements as prompts. Ground-truth answers are used only for evaluation and for the GT-Reward baseline. 

In mathematical reasoning task, we train Qwen2.5-3B and Qwen2.5-7B Base models on MATH dataset~\citep{math7500}. Given a prompt ${x}$, we sample a group of $G{=}8$ completions. For the GRPO baseline, we extract the final answer from each completion, compare it with the reference answer and assign a binary reward ($1$ for correct and $0$ for incorrect after standard normalization). For our method, we also sample a group of $G=8$ completions, but we calculate the reward $R_{\text{GN}}$ for each completion without any external verifier. We then compute the group-relative advantages from these rewards and update the policy following the same GRPO optimization procedure.

In code generation task, following the setting of INTUITOR~\citep{intuitor}, we only train Qwen2.5-3B Base models on the first 3,200 problems of the CodeContests dataset~\citep{alphacode} and adopt the same optimization strategy in the mathematical reasoning task.

More detailed implementation details are provided in Appendix~\ref{sec:additional_training_details}.

\subsection{Main Results}
\begin{figure*}[t]
  \centering
  \begin{subfigure}[t]{0.96\columnwidth}
    \centering
    \includegraphics[width=\linewidth]{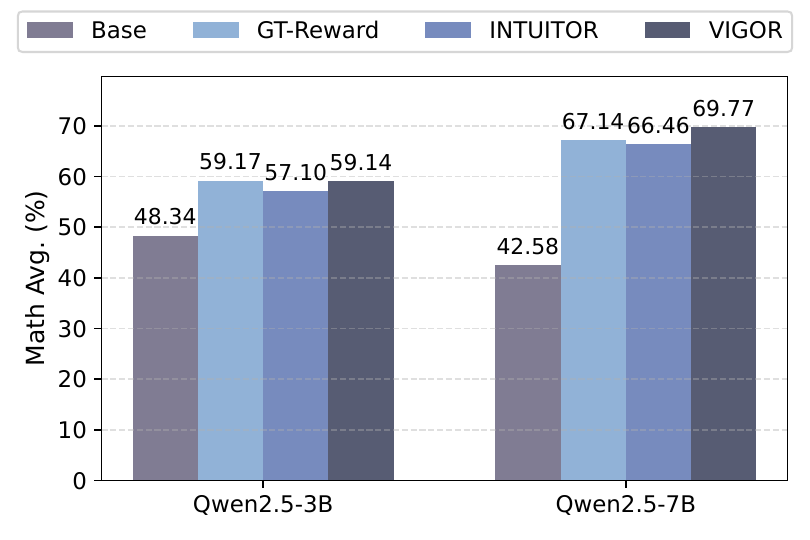}
    \caption{Average performance on math benchmarks.}
  \end{subfigure}
  \hfill
  \begin{subfigure}[t]{0.96\columnwidth}
    \centering
    \includegraphics[width=\linewidth]{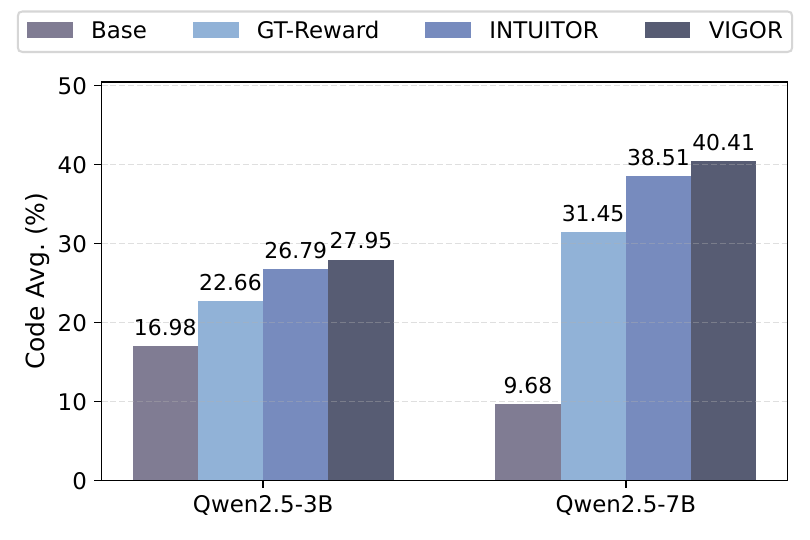}
    \caption{Average performance on code benchmarks.}
  \end{subfigure}
  \caption{Average benchmark performance across model scales.}
  \label{fig:avg_summary}
\end{figure*}
Figure~\ref{fig:avg_summary} provides an overview of average math and code performance across model scales, comparing Base, GRPO, INTUITOR, and VIGOR. The detailed experimental results are reported in Table~\ref{tab:training_on_math} and Table~\ref{tab:training_on_codecontests}. These results show that our method consistently improves the reasoning ability of Qwen2.5 Base model on both mathematics and code tasks.
\begin{figure*}[t]
  \centering

  \begin{subfigure}[t]{0.24\textwidth}
    \centering
    \includegraphics[width=\linewidth]{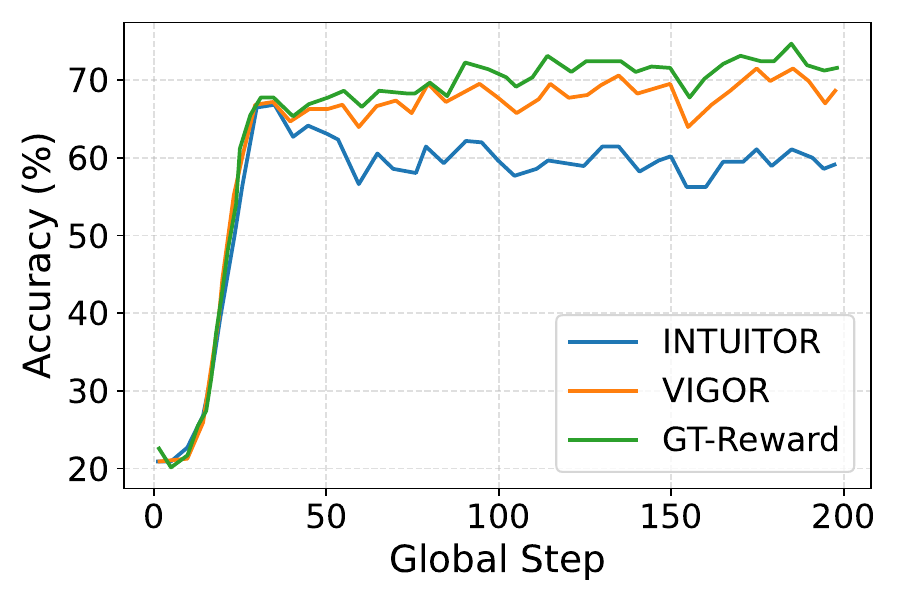}
    \caption{Accuracy}\label{fig:acc}
  \end{subfigure}\hfill
  \begin{subfigure}[t]{0.24\textwidth}
    \centering
    \includegraphics[width=\linewidth]{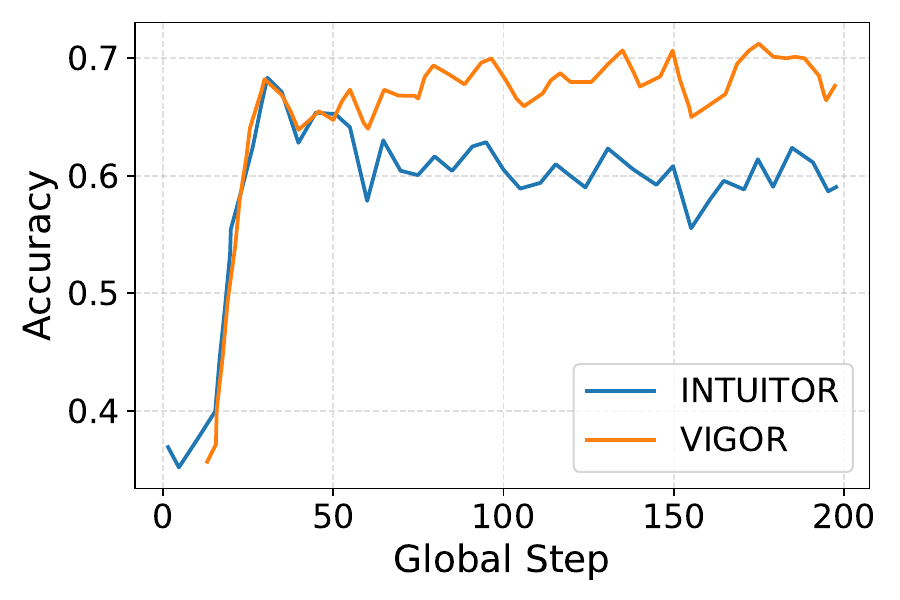}
    \caption{Top-25\% accuracy}\label{fig:top25}
  \end{subfigure}\hfill
  \begin{subfigure}[t]{0.24\textwidth}
    \centering
    \includegraphics[width=\linewidth]{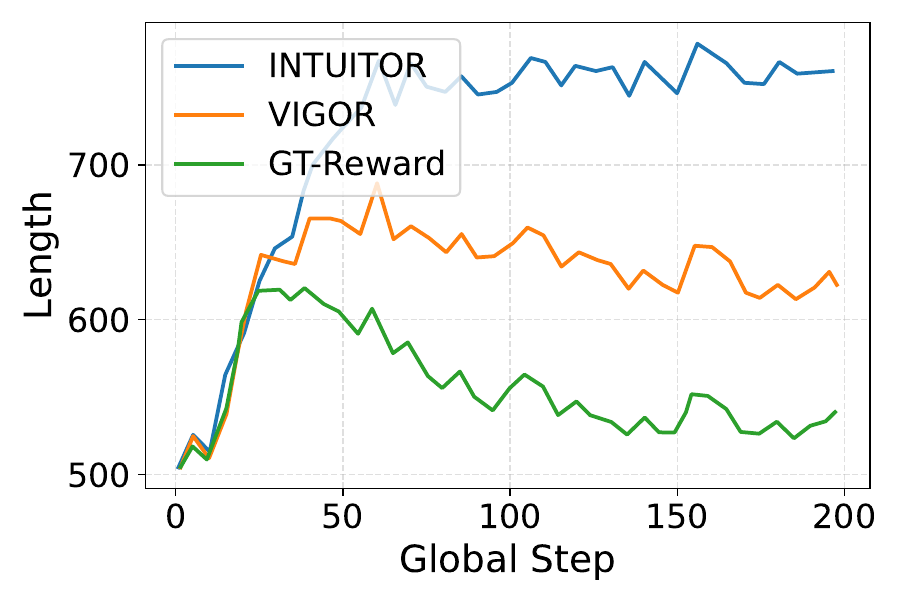}
    \caption{Length}\label{fig:len}
  \end{subfigure}\hfill
  \begin{subfigure}[t]{0.24\textwidth}
    \centering
    \includegraphics[width=\linewidth]{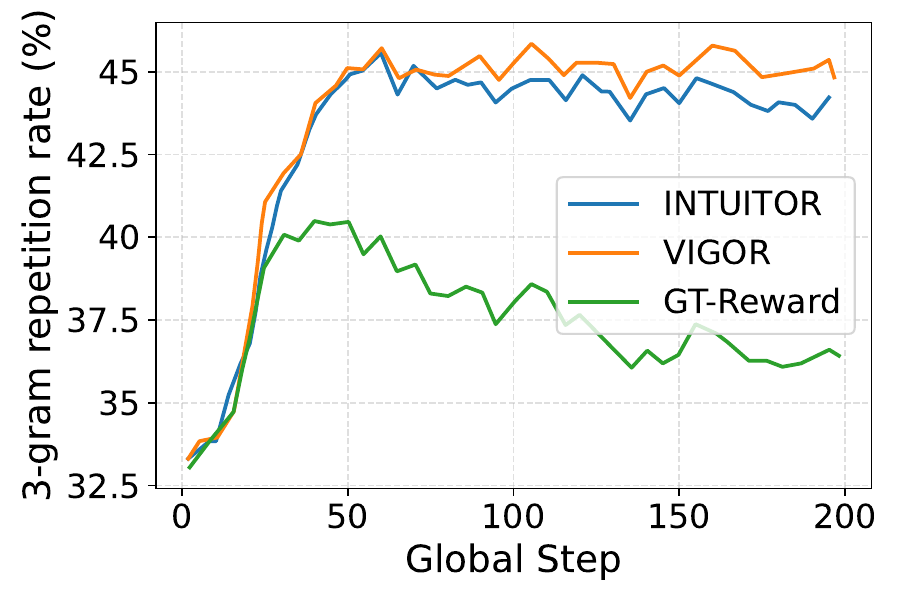}
    \caption{3-gram repetition}\label{fig:repeat}
  \end{subfigure}

  \caption{Training dynamics under verifier-free intrinsic rewards.
  We track 198 training steps of Qwen2.5-7B-Base optimized with INTUITOR (self-certainty reward) and VIGOR (gradient-norm reward), and compare against a verifiable GT-Reward baseline.
  We report (a) overall accuracy, (b) accuracy of the top-25\% completions ranked by within-group reward, (c) average completion length, and (d) mean 3-gram repetition rate.
  We omit GT-Reward in (b) since its exact-match reward directly encodes correctness.}
  \label{fig:training_dynamics}
\end{figure*}

\begin{table*}[t]
    \centering
    \small
    \setlength{\tabcolsep}{3.5pt}
    \renewcommand{\arraystretch}{1.05}
    \begin{tabular}{@{}lccccccc cc@{}}
        \toprule
         &
        \multicolumn{4}{c}{\textbf{Mathematics}} &
        \multicolumn{3}{c}{\textbf{Code}} &
        \multicolumn{1}{c}{\textbf{Multi-task}} &
        \multicolumn{1}{c}{\textbf{Instruction}} \\
        \cmidrule{1-10}
        \textbf{Methods} &
        \textbf{GSM8K} & \textbf{MATH500} & \textbf{AMC} & \textbf{Avg.} &
        \textbf{LiveCodeBench} & \textbf{CRUX} & \textbf{Avg.} &
        \textbf{MMLU-Pro} & \textbf{IFEval} \\
        \midrule

        \multicolumn{10}{@{}l}{\textbf{Qwen2.5-3B-Base}} \\
        \addlinespace[2pt]
        - \textbf{Full (VIGOR)}     & \textbf{81.80}   & \textbf{64.60}   & \textbf{31.02}   & \textbf{59.14}    & \textbf{15.90}   & \textbf{40.00}   & \textbf{27.95}   & 32.65   & \textbf{31.72}   \\
        \quad w/o $\sqrt{T}$   & 0.08 & 60.40 & 1.66 & 20.71 & 0.00    & 0.00    & 0.00 & \textbf{36.39}   & 29.30   \\
        \quad w/o rank           & 81.35   & 63.00   & 29.66   & 58.00    & 15.26   & 38.88   &  27.07  & 33.44   & 30.19   \\

        \addlinespace[0.5em]
        \multicolumn{10}{@{}l}{\textbf{Qwen2.5-7B-Base}} \\
        \addlinespace[2pt]
        - \textbf{Full (VIGOR)}     &
          \textbf{88.70} & \textbf{76.20} & \textbf{44.42} & \textbf{69.77} &
          24.45 & 56.38 & 40.42 &
          \textbf{43.09} & 37.03 \\
        \quad w/o $\sqrt{T}$    &
          87.87 & 75.00 & 42.01 & 68.29 &
          24.92 & \textbf{57.75} & \textbf{41.34} &
          41.34 & 37.23 \\
        \quad w/o rank           &
          88.32 & 75.20 & 44.12 & 69.21 &
          \textbf{25.50} & 54.62 & 40.06 &
          34.19 & \textbf{38.32} \\

        \bottomrule
    \end{tabular}
    \caption{Ablation study of the two key components in VIGOR: $\sqrt{T}$ length correction for gradient norms and rank-based normalization for intrinsic signal. Avg. is the arithmetic mean over tasks in each block. \textbf{Bold} indicates the best result within each model block.}
    \label{tab:ablation}
\end{table*}

\paragraph{Training on MATH.}
Table~\ref{tab:training_on_math} reports the experimental results trained on MATH. VIGOR substantially improves Qwen2.5 Base models on mathematic benchmarks, with the average performance increasing from 48.34\% to 59.14\% on Qwen2.5-3B and from 42.58\% to 69.77\% on Qwen2.5-7B. Specifically, VIGOR improves GSM8K and AMC by +45.64\% and +22.74\% on Qwen2.5-7B, indicating that the intrinsic gradient-norm signal can effectively drive reasoning improvement without answer supervision. Beyond in-domain tasks, VIGOR also transfers to code when trained on MATH, improving the average accuracy of code benchmarks by +10.97\% on Qwen2.5-3B and +30.73\% on Qwen2.5-7B. In terms of general multi-task and instruction-following ability, VIGOR largely preserves performance on MMLU-Pro and IFEval while outperforming the verifier-free baseline.

\paragraph{Training on CodeContests.}
Table~\ref{tab:training_on_codecontests} reports results of training on CodeContests with Qwen2.5-3B-Base. Since our main focus is mathematical reasoning (where we run full experiments on both 3B and 7B), we include CodeContests as a lightweight sanity check to verify that our verifier-free reward remains effective for prompt-only code generation training. For simplicity and controlled training cost, we only use a small subset of 3,200 problems from the CodeContests training split, aiming to validate generality rather than to maximize code performance.

Despite this lightweight setup, VIGOR still yields gains on code reasoning, improving the average code performance from 16.98\% to 23.64\%. In addition, CodeContests training also transfers to mathematical reasoning: VIGOR improves the math average from 48.34\% to 56.57\%, with consistent gains on GSM8K, MATH500, and AMC.
We note, however, that on Code Avg., VIGOR (23.64\%) underperforms INTUITOR (25.43\%), particularly on CRUX. We attribute this to the nature of competitive programming, where correctness often hinges on discrete algorithmic choices that may not be well-reflected in gradient-norm smoothness.

\subsection{Analysis}
To better understand the training dynamics of our proposed gradient-norm reward, we visualize various metrics during training in Figure~\ref{fig:training_dynamics}. We compare VIGOR against two baselines: (i) INTUITOR, a verifier-free method that uses self-certainty as the reward, and (ii) GT-Reward, which applies GRPO with an exact-match reward against the reference answer. In Figure~\ref{fig:training_dynamics}, accuracy denotes training-time evaluation accuracy computed by exact match against references, used only for monitoring.

\paragraph{Training Dynamics and Stability.}
As shown in Figure~\ref{fig:training_dynamics}(\subref{fig:acc}), all methods improve rapidly during the initial stage of training. However, the trajectories diverge in later updates: INTUITOR exhibits a clear late-stage regression, where accuracy drifts downward rather than plateauing. This pattern is consistent with reward-proxy degeneration—as the policy adapts, confidence-based internal feedback becomes increasingly exploitable and can be optimized without improving correctness, causing the update direction to gradually decouple from the true objective. In contrast, VIGOR maintains a stable improvement and consistently achieves higher accuracy, indicating that the gradient-norm reward remains a more reliable training signal throughout optimization.

We hypothesize that this stability stems from defining the reward in the model’s parameter space: the gradient norm aggregates signals over a high-dimensional set of parameters and is therefore less sensitive to token-level distribution shifts. In contrast, entropy- or confidence-based rewards are computed from the next-token distribution over the vocabulary and can be more easily exploited by superficial behaviors.
\begin{figure*}[t]
  \centering

  \begin{subfigure}{0.24\textwidth}
    \centering
    \includegraphics[width=\linewidth]{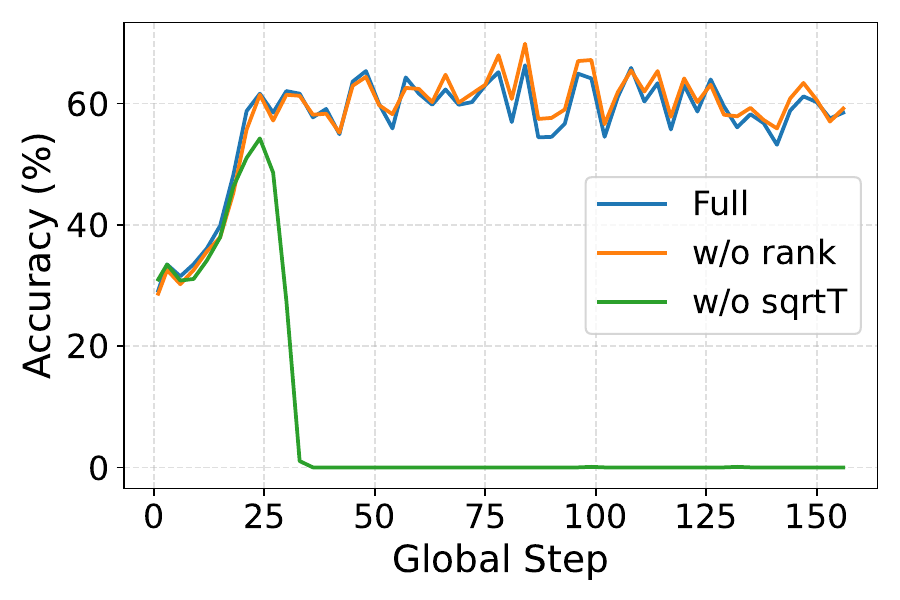}
    \caption{3B Accuracy}
    \label{fig:abl-acc-3b}
  \end{subfigure}\hfill
  \begin{subfigure}{0.24\textwidth}
    \centering
    \includegraphics[width=\linewidth]{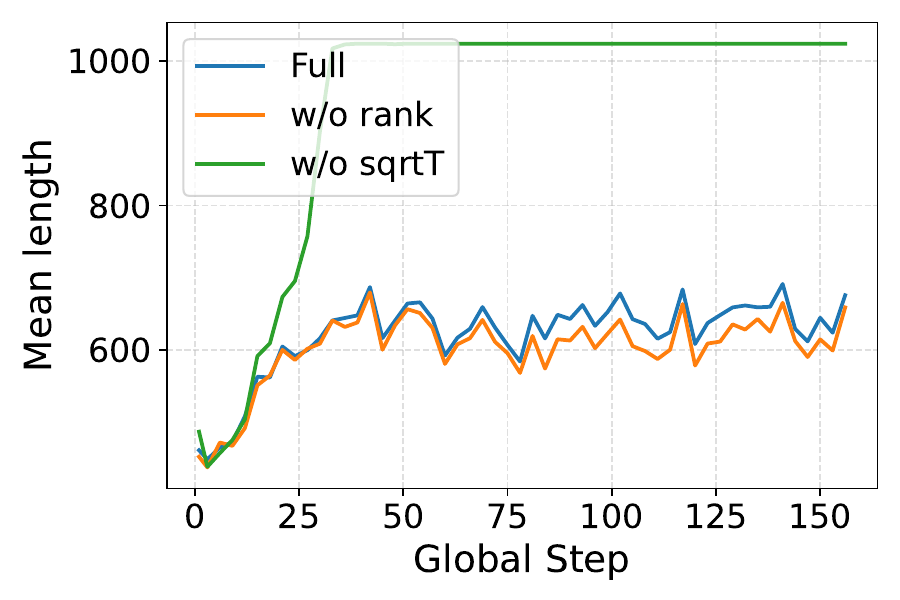}
    \caption{3B Mean length}
    \label{fig:abl-len-3b}
  \end{subfigure}\hfill
  \begin{subfigure}{0.24\textwidth}
    \centering
    \includegraphics[width=\linewidth]{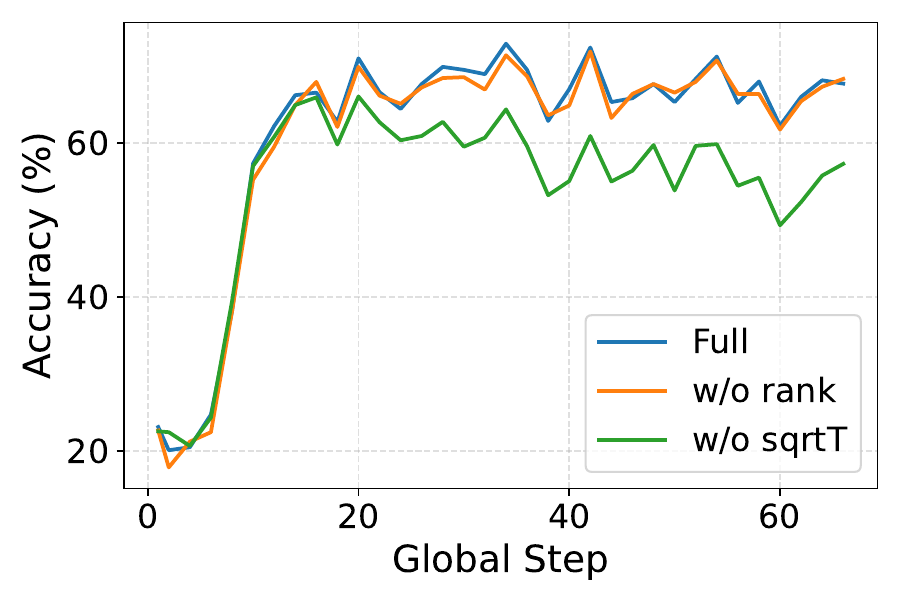}
    \caption{7B Accuracy}
    \label{fig:abl-acc-7b}
  \end{subfigure}\hfill
  \begin{subfigure}{0.24\textwidth}
    \centering
    \includegraphics[width=\linewidth]{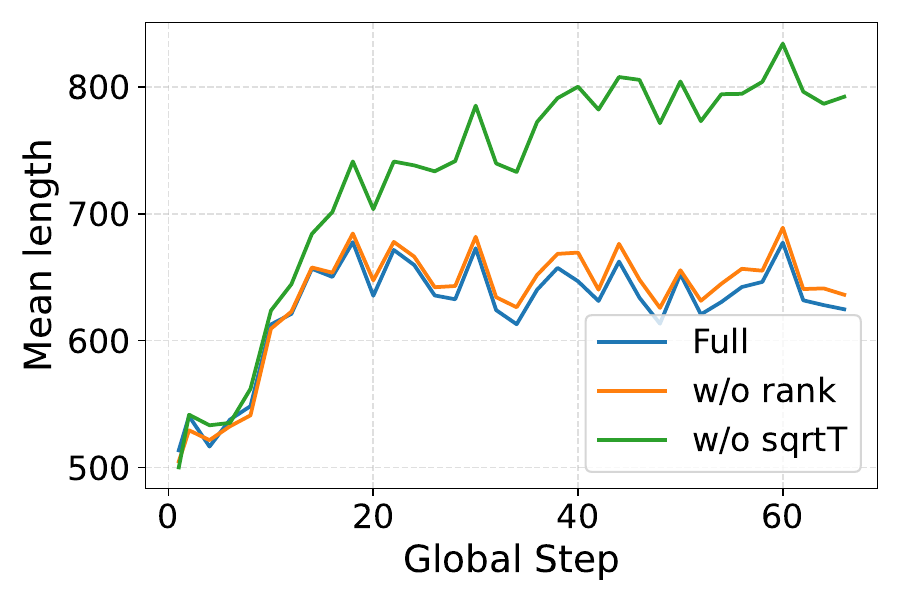}
    \caption{7B Mean length}
    \label{fig:abl-len-7b}
  \end{subfigure}

  \caption{Training dynamics of the Qwen2.5-3B and 7B model under ablations.
  \textit{Full} uses the complete \textsc{VIGOR} reward; \textit{w/o rank} removes within-group rank-based normalization; \textit{w/o $\sqrt{T}$} removes length correction.
  Removing $\sqrt{T}$ induces severe length-bias exploitation: generations become increasingly long and the accuracy collapses (especially on 3B).}
  \label{fig:ablation_fig}
\end{figure*}

\paragraph{Reward Reliability.}
Figure~\ref{fig:training_dynamics}(\subref{fig:top25}) reports the accuracy of the top-25\% completions ranked by the within-group intrinsic reward (with group size $G{=}8$, this corresponds to the top-2 completions per prompt), assessing whether high intrinsic rewards consistently align with correctness.
VIGOR maintains higher and relatively stable top-25\% accuracy across training, exhibiting the reliability of its intrinsic reward signal.
In contrast, INTUITOR exhibits a clear degradation in top-25\% accuracy in later stages, implying that high-reward samples become progressively less aligned with correctness. This behavior is consistent with reward degeneration: as training proceeds, confidence-based internal signals can be increasingly exploited.
We omit GT-Reward in Figure~\ref{fig:training_dynamics}(\subref{fig:top25}) since exact-match rewards are tightly coupled with correctness, making this analysis less informative.

\begin{table}[t]
\centering
\small
\begin{tabular}{@{}ccc@{}}
\toprule
Rank & Step 10 Acc.\,(\%) & Step 20 Acc.\,(\%) \\
\midrule
1 (best)  & 70.50 & 72.30 \\
2          & 68.20 & 71.10 \\
3          & 67.90 & 69.80 \\
4          & 67.00 & 67.40 \\
5          & 66.10 & 67.10 \\
6          & 60.70 & 66.00 \\
7          & 55.30 & 65.00 \\
8 (worst) & 52.70 & 63.40 \\
\bottomrule
\end{tabular}
\caption{Rank–accuracy monotonicity on Qwen2.5-3B-Base trained on MATH. For each prompt we sample $G{=}8$ completions, rank them by the length-corrected gradient-norm score $\sqrt{T}\lVert\mathbf{g}\rVert_2$ (rank 1 = lowest norm, best), and report the average accuracy per rank position across all prompts.}
\label{tab:rank_accuracy}
\end{table}
 
\paragraph{Rank--Accuracy Monotonicity.}
To quantify how well the gradient-norm ranking aligns with task correctness, we conduct a fine-grained monotonicity analysis on Qwen2.5-3B-Base trained on MATH. At training steps 10 and 20, for each prompt we sample $G{=}8$ completions, compute the length-corrected score $\sqrt{T}\lVert\mathbf{g}\rVert_2$, rank the 8 completions within each prompt (ascending; lower score corresponds to better rank), and report the average accuracy per rank position across all prompts.
As shown in Table~\ref{tab:rank_accuracy}, a clear monotonic trend holds at both checkpoints: completions ranked better by gradient norm consistently achieve higher accuracy. At step~10, accuracy ranges from 70.50\% (rank~1) to 52.70\% (rank~8), yielding a gap of 17.8 percentage points. At step~20, the gap narrows to 8.9 points as overall accuracy improves and the model approaches convergence, which is expected behavior. Importantly, the monotonic ordering is preserved throughout, confirming that the gradient-norm ranking remains a directionally consistent signal rather than degrading over training. This fine-grained analysis complements the top-25\% accuracy trend in Figure~\ref{fig:training_dynamics}(\subref{fig:top25}) and provides statistical validation that the intrinsic gradient-norm reward is meaningfully correlated with correctness.

\paragraph{Output Length and Repetition.}
Figure~\ref{fig:training_dynamics}(\subref{fig:len}) shows clear differences in generation length. INTUITOR exhibits a pronounced increase in completion length over training, while VIGOR produces substantially shorter outputs; GT-Reward yields the shortest generations overall.
We further examine surface-level degeneration using the mean 3-gram repetition rate, which measures the fraction of repeated 3-grams within a completion.
As shown in Figure~\ref{fig:training_dynamics}(\subref{fig:repeat}), VIGOR exhibits repetition rates comparable to INTUITOR, and both are higher than GT-Reward. This observation decouples length growth from repetition: it suggests that VIGOR's gains are driven by curbing the tendency for unconstrained length inflation, ensuring optimization focuses on reasoning quality rather than length hacking.



\subsection{Ablation Study}
\label{sec:ablation}

To investigate the contribution of each component in VIGOR, we conduct ablation studies on Qwen2.5-3B and 7B Base model trained on MATH dataset. The results are reported in Table~\ref{tab:ablation}.

\paragraph{Impact of $\sqrt{T}$ length correction.}
The $\sqrt{T}$ length correction  proves crucial for stabilizing the training. As shown in Figure~\ref{fig:ablation_fig}(\subref{fig:abl-acc-3b}) and~\ref{fig:ablation_fig}(\subref{fig:abl-len-3b}), removing this length correction causes catastrophic degradation in Qwen2.5-3B-Base, resulting in near-zero accuracy on GSM8K/AMC, a complete loss of transferability to code tasks and the emergence of length hacking.
This highlights that the raw gradient norm of token-averaged NLL exhibits a strong length-dependent bias, making the intrinsic signal exploitable and unstable without correction.
In contrast, the drop on 7B is smaller, suggesting larger models may provide a less noisy intrinsic signal, but $\sqrt{T}$ correction remains beneficial overall.
\paragraph{Impact of within-group rank shaping.}
We observe that the necessity of rank shaping correlates with model scale. For the Qwen2.5-3B Base model, the removal of rank shaping results in a trade-off on general tasks, with slight gains on MMLU-Pro offset by losses on IFEval. In contrast, the Qwen2.5-7B Base model relies on it to maintain general capabilities. Notably, removing the rank shaping causes a severe accuracy regression of 8.90\% on MMLU-Pro. We attribute this to the instability of the raw gradient-norm reward across prompts—its scale varies substantially and occasionally contains extreme outliers. Rank shaping mitigates this issue by using only within-prompt relative order, preventing a small subset of high-magnitude cases from dominating the updates and steering training away from general knowledge.

Additional ablation results, including normalization variants and a scalable LM-head-only variant, are provided in Appendix~\ref{sec:additional_ablations}.

\section{Conclusions}
In this work, we introduce VIGOR, a verifier-free intrinsic reward for RL post-training of LLMs when gold answers or domain verifiers are unavailable. VIGOR uses per-completion gradient norms as reward, favoring low-norm completions for stable optimization, with length correction and within-group rank shaping for robustness. Across math and code reasoning benchmarks, VIGOR improves performance and stabilizes training. We hope VIGOR motivates further work on scalable verifier-free RL post-training.

\section*{Limitations}
Our study primarily targets verifiable reasoning tasks; it remains unclear how well VIGOR transfers to more open-ended generation settings (e.g., long-form writing or dialogue), where additional constraints may be needed to avoid pathological optimization. In addition, VIGOR requires per-sample gradient-norm computation, which introduces non-trivial automatic-differentiation overhead compared to forward-only signals such as likelihood or entropy, and although our LM-head-only variant largely mitigates this cost, scaling to substantially larger models may still require further approximations. More broadly, gradient norm is only a proxy objective and may not consistently track downstream utility, leaving room for potential misalignment or reward exploitation.

\section*{Acknowledgments}
This work was supported in part by the ``Pioneer'' and ``Leading Goose'' R\&D Program of Zhejiang (No.\ 2025C02032).
This work was also supported by Ant Group through the CAAI-Ant Research Fund and the Earth System Big Data Platform of the School of Earth Sciences, Zhejiang University.

\bibliography{custom}

\appendix

\begin{table*}[t]
  \centering
  \small
  \setlength{\tabcolsep}{6pt}
  \renewcommand{\arraystretch}{1.12}
  \begin{tabular}{@{}lcc@{}}
    \toprule
    \textbf{Hyperparameter} & \textbf{Qwen2.5-3B-Base} & \textbf{Qwen2.5-7B-Base} \\
    \midrule
    Batch size per GPU & 4 & 4 \\
    Gradient Accumulation & 32 & 32 \\
    Max prompt length & 512 & 512 \\
    Max response length & 1024 & 1024 \\
    Train steps & 156 & 66 \\
    Learning rate & 2e{-}6 & 1e{-}6 \\
    Rollouts $G$ & 8 & 8 \\
    Clip ratio & 0.2 & 0.2 \\
    Warmup schedule & Cosine & Cosine \\
    Warmup ratio & 0.1 & 0.1 \\
    KL coefficient $\beta$ & 0.01 & 0.01 \\
    Optimizer & \multicolumn{2}{c}{AdamW ($\beta_1{=}0.9, \beta_2{=}0.999, \epsilon{=}10^{-8}$)} \\
    Training temperature & 0.9 & 0.9 \\
    \bottomrule
  \end{tabular}
  \caption{Training configuration on \textbf{MATH}. All methods (GT-Reward, INTUITOR, and VIGOR) share the same configuration; they differ only in reward computation (Table~\ref{tab:reward_def}).}
  \label{tab:training_config_math}
\end{table*}

\begin{table*}[t]
\centering
\small
\begin{tabular}{lccccc}
\toprule
Method & Steps & Wall-clock & Min/step & GPU-hours & Avg. mem (GB) \\
\midrule
GT-Reward              & 66 & 2h34m08s & 2.34 & 20.55 & 66.23 \\
INTUITOR               & 66 & 3h57m35s & 3.60 & 31.68 & 66.32 \\
VIGOR                  & 66 & 3h35m00s & 3.26 & 28.67 & 74.00 \\
VIGOR (LM-head-only)   & 66 & 2h47m16s & 2.53 & 22.31 & 66.25 \\
\bottomrule
\end{tabular}
\caption{End-to-end training cost for Qwen2.5-7B on 8$\times$H800 (7 training GPUs + 1 vLLM rollout GPU) for 66 optimization steps. Wall-clock includes both vLLM rollouts and GRPO optimization. GPU-hours are computed as $8 \times$ wall-clock hours. Avg. mem reports the average GPU memory usage measured by \texttt{nvidia-smi}.}
\label{tab:budget_7b_h800}
\end{table*}


\begin{table*}[t]
  \centering
  \small
  \setlength{\tabcolsep}{6pt}
  \renewcommand{\arraystretch}{1.12}
  \begin{tabular}{@{}lc@{}}
    \toprule
    \textbf{Hyperparameter} & \textbf{Qwen2.5-3B-Base} \\
    \midrule
    Batch size per GPU & 4 \\
    Gradient Accumulation & 32 \\
    Max prompt length & 1024 \\
    Max response length & 2048 \\
    Train steps & 66 \\
    Learning rate & 5e{-}6 \\
    Rollouts $G$ & 8 \\
    Clip ratio & 0.2 \\
    Warmup schedule & Cosine \\
    Warmup ratio & 0.1 \\
    KL coefficient $\beta$ & 0.01 \\
    Optimizer & AdamW ($\beta_1{=}0.9, \beta_2{=}0.999, \epsilon{=}10^{-8}$) \\
    Training temperature & 0.9 \\
    \bottomrule
  \end{tabular}
  \caption{Training configuration on \textbf{CodeContests} with Qwen2.5-3B-Base. All methods (GT-Reward, INTUITOR, and VIGOR) share the same configuration; they differ only in reward computation (Table~\ref{tab:reward_def}).}
  \label{tab:training_config_code}
\end{table*}


\begin{table*}[t]
  \small
  \centering
  \setlength{\tabcolsep}{6pt}
  \renewcommand{\arraystretch}{1.12}
  \begin{tabular}{@{}l p{0.60\textwidth} p{0.22\textwidth}@{}}
    \toprule
    \textbf{Method} &
    \textbf{Reward signal} &
    \textbf{External requirement} \\
    \midrule

    GT-Reward &
    Outcome reward with an exact-match verifier. &
    Gold answers / task-specific verifier. \\

    INTUITOR &
    Self-certainty reward computed from the policy model's likelihood signal. &
    None. \\

    VIGOR (Ours) &
    Gradient-norm intrinsic reward with length correction and rank shaping. &
    None. \\

    \bottomrule
  \end{tabular}
  \caption{Reward signals used for GRPO training. Within each setting, all hyperparameters are identical across methods; the only difference lies in reward computation.}
  \label{tab:reward_def}
\end{table*}

\begin{table}
  \small
  \centering
  \setlength{\tabcolsep}{4pt}
  \renewcommand{\arraystretch}{1.05}
  \begin{tabular}{lrl}
    \toprule
    \textbf{Dataset} & \textbf{Problems} & \textbf{Usage} \\
    \midrule
    MATH500 & 500 & Math eval \\
    GSM8K & 1{,}319 & Math eval \\
    AMC & 83 & Math eval \\
    LiveCodeBench & 1{,}055 & Code eval \\
    CRUX~ & 800 & Code eval \\
    IFEval & 541 & Instruction eval \\
    MMLU-Pro & 12{,}032 & Multi-task eval \\
    \bottomrule
  \end{tabular}
  \caption{Datasets used for evaluation.}
  \label{tab:datasets}
\end{table}
\section{Additional Training Details}
\label{sec:additional_training_details}
We provide additional details of the training setup and hyperparameters.

\subsection{Training Setting}
We conduct GRPO training using the Open-R1 framework and its vLLM-based rollout backend for efficient generation~\citep{openr1}.

In our setup, we reserve one GPU for vLLM rollout generation. Therefore, for post-training on \textbf{MATH}, Qwen2.5-3B-Base trained on 4$\times$A6000 uses 3 training GPUs (3 training processes) and runs 156 optimization steps, while Qwen2.5-7B-Base trained on 8$\times$H800 uses 7 training GPUs (7 training processes) and runs 66 optimization steps. The step counts are chosen according to the available compute budget under each hardware setting.
For \textbf{CodeContests}, we also train on 4$\times$A6000 (with one GPU reserved for rollout), but use only a 3,200-problem subset and run 66 optimization steps.

\paragraph{Post-training on MATH (main setting).}
Our main experiments focus on mathematical reasoning. We run full post-training on MATH with both Qwen2.5-3B-Base and Qwen2.5-7B-Base.
Table~\ref{tab:training_config_math} summarizes the training configuration for this setting.
For a fair comparison, we keep all hyperparameters identical across methods within this setting, and the only difference lies in reward computation (Table~\ref{tab:reward_def}).

\paragraph{Post-training on CodeContests (lightweight sanity check).}
We additionally include CodeContests as a lightweight sanity check to verify that our verifier-free reward remains effective for prompt-only code generation training.
To keep training cost controlled, we only post-train Qwen2.5-3B-Base and use a small subset of 3,200 problems from the CodeContests training split.
In this setting, we compare \emph{only} two verifier-free methods, INTUITOR and VIGOR, and do not include execution-based RLVR baselines that require task-specific verifiers.
Table~\ref{tab:training_config_code} summarizes the training configuration for this setting.
For a fair comparison, we keep all hyperparameters identical across methods within this setting, and the only difference lies in reward computation (Table~\ref{tab:reward_def}).

\paragraph{Gradient-norm computation.}
We compute the gradient norm sequentially across the $G$ completions within each prompt. For each completion $y_i$, we construct the scalar loss $\ell_{\text{mean}}(x,y_i)$ and call \texttt{torch.autograd.grad} to obtain $\nabla_\theta \ell_{\text{mean}}$, from which we compute $\|\mathbf g\|_2$. This requires $G{=}8$ sequential backward passes per prompt. We do not retain computational graphs across completions to control memory usage.

\subsection{Training Budget and Resource Usage}
We report end-to-end wall-clock time including both vLLM rollouts and GRPO optimization, as well as average GPU memory usage measured by \texttt{nvidia-smi}. Table~\ref{tab:budget_7b_h800} provides a representative example for Qwen2.5-7B trained on 8$\times$H800 for 66 optimization steps.

Compared to GT-Reward, INTUITOR and VIGOR incur additional training cost, taking 3h57m and 3h35m versus 2h34m, respectively.
In terms of GPU memory, VIGOR requires higher maximum reserved memory across GPUs (74.0\,GB), compared to 66.2--66.3\,GB for GT-Reward and INTUITOR. 
We attribute the higher memory consumption of VIGOR to computing gradient-norm based intrinsic rewards, which requires additional gradient statistics during reward computation. 
Overall, VIGOR trains slightly faster than INTUITOR, trading off higher GPU memory usage.
The LM-head-only variant further reduces wall-clock time to 2h47m and memory usage to 66.25\,GB, approaching the cost of GT-Reward.

\subsection{Prompt Format}
\paragraph{MATH prompts format.}
We use the Qwen chat template to construct prompts, following a \texttt{system}--\texttt{user}--\texttt{assistant} format. We use different system prompts for Qwen2.5-3B-Base and Qwen2.5-7B-Base on MATH.
For the 3B model, we apply a more constrained instruction to better regularize the smaller model:
\texttt{``Let's think step by step and output the final answer within \textbackslash boxed\{\}.''}
For the 7B model, we use a more general assistant-style instruction:
\texttt{``You are a helpful AI Assistant, designed to provide well-reasoned and detailed responses. Please provide a step-by-step solution to the following problem.''} 

The following shows a concrete example used during 7B training:
\smallskip

\noindent{\footnotesize\ttfamily
<|im\_start|>system\\
You are a helpful AI Assistant, designed to provide well-reasoned and detailed responses.\\
Please provide a step-by-step solution to the following problem.<|im\_end|>\\
<|im\_start|>user\\
Let $f(x)=3x^2-7$ and $g(f(4))=9$. What is $g(f(-4))$?<|im\_end|>\\
<|im\_start|>assistant\\
}

\paragraph{CodeContests prompt format.}

For CodeContests, we only post-train Qwen2.5-3B-Base.
We construct prompts using the same Qwen chat template in a \texttt{system}--\texttt{user}--\texttt{assistant} format and start generation from the final \texttt{<|im\_start|>assistant} marker.
The system prompt is designed to elicit code-only outputs:
\texttt{``You are an AI designed to help solve competitive programming problems by generating Python code.''}
The following shows a concrete example used during 3B training:

\smallskip
\noindent{\footnotesize\ttfamily
<|im\_start|>system\\
You are a competitive programming assistant. Write a correct and efficient Python 3 program that reads from stdin and writes to stdout. Output only the code.<|im\_end|>\\
<|im\_start|>user\\
Write a Python function or program that fulfills the task described below. Provide only the code as output.\\
Task Description: Given three integers $A_1, A_2, A_3$, print \texttt{bust} if $A_1{+}A_2{+}A_3 \ge 22$, otherwise print \texttt{win}.\\
Input: $A_1\ A_2\ A_3$ \quad Output: \texttt{bust} or \texttt{win}.\\
{[}\dots{]}\\
<|im\_start|>assistant\\
}

\section{Additional Evaluation Details}
\label{sec:additional_evaluation_details}

\paragraph{Evaluation Setting.}
We evaluate the model on the datasets listed in Table~\ref{tab:datasets} using the corresponding evaluation scripts.
For all benchmarks except AMC, we use greedy decoding with temperature = 0 and generate one completion per prompt (pass@1).
For AMC, since it contains only 83 problems (Table~\ref{tab:datasets}), the evaluation variance can be higher; therefore we use temperature = 0.6 and report avg@8 (average over 8 independent samples/runs) for a more stable estimate.
\paragraph{Metrics.} We report pass@1 for math and code generation benchmarks (MATH500, GSM8K, LiveCodeBench-v6, CRUX) and accuracy for multiple-choice benchmarks (MMLU-Pro). Instruction following is evaluated with the IFEval metric as defined in its benchmark.

\section{Additional Ablation Results}
\label{sec:additional_ablations}
\begin{table}[t]
\centering
\small

\begin{tabular}{@{}lcccc@{}}
\toprule
Method & GSM8K & MATH500 & AMC & Avg. \\
\midrule
INTUITOR            & 87.19 & 76.20 & 35.99 & 66.46 \\
INTUITOR w/ rank    & 88.93 & 74.60 & 40.36 & 67.96 \\
VIGOR               & 88.70 & 76.20 & 44.42 & 69.77 \\
\bottomrule
\end{tabular}
\caption{Disentangling regularization from the gradient-norm signal on Qwen2.5-7B trained on MATH.}
\label{tab:ablation_rank_transfer}
\end{table}
\begin{table}[t]
\centering
\small
\begin{tabular}{@{}lcccc@{}}
\toprule
Method & GSM8K & MATH500 & AMC & Avg. \\
\midrule
VIGOR (Rank)    & 88.70 & 76.20 & 44.42 & 69.77 \\
VIGOR (Min-Max) & 88.70 & 74.40 & 43.97 & 69.02 \\
\bottomrule
\end{tabular}
\caption{Rank vs.\ Min-Max normalization on Qwen2.5-7B trained on MATH.}
\label{tab:ablation_norm}
\end{table}
\begin{table}[t]
\centering
\small
\begin{tabular}{@{}lcccc@{}}
\toprule
Variant & GSM8K & MATH500 & AMC & Avg. \\
\midrule
Full model   & 88.70 & 76.20 & 44.42 & 69.77 \\
LM-head only & 88.63 & 75.40 & 44.57 & 69.53 \\
\bottomrule
\end{tabular}
\caption{LM-head-only gradient-norm variant on Qwen2.5-7B trained on MATH.}
\label{tab:ablation_lmhead}
\end{table}
We report additional ablation experiments on Qwen2.5-7B-Base trained on MATH to complement the main ablation study in Section~\ref{sec:ablation}. All variants use the same hyperparameters and differ only in the specified component.

\paragraph{Disentangling regularization from the gradient-norm signal.}
To verify that VIGOR's improvements are not solely attributable to rank normalization, we apply the same rank-based shaping to INTUITOR's confidence reward. As shown in Table~\ref{tab:ablation_rank_transfer}, INTUITOR with rank normalization improves from 66.46\% to 67.96\% Math Avg., but remains below VIGOR (69.77\%), confirming that the gradient-norm signal itself is the primary driver of improvement.

\paragraph{Rank vs.\ Min-Max normalization.}
We compare rank normalization against Min-Max normalization, which maps rewards linearly to $[-1,+1]$ while preserving within-group magnitude differences. Table~\ref{tab:ablation_norm} shows that rank normalization slightly outperforms Min-Max (69.77\% vs.\ 69.02\%), suggesting that discarding magnitude information in favor of bounded, outlier-robust updates is preferable for the high-variance gradient-norm reward.

\paragraph{LM-head-only gradient norm.}
To reduce computational overhead, we restrict gradient-norm computation to the LM head parameters only. As shown in Table~\ref{tab:ablation_lmhead}, this variant achieves comparable performance (69.53\% vs.\ 69.77\% Math Avg.) while substantially reducing memory and wall-clock cost (Table~\ref{tab:budget_7b_h800}), making VIGOR practical for larger models.

\end{document}